\pdfoutput=1

\documentclass[11pt]{article}

\usepackage{ACL2023}

\usepackage{times}
\usepackage{latexsym}

\usepackage[T1]{fontenc}

\usepackage[utf8]{inputenc}

\usepackage{microtype}

\usepackage{inconsolata}

\usepackage{booktabs}
\usepackage{url}

\title{Extracting Psychological Indicators Using Question Answering}

\author{Luka Pavlović \\
  Faculty of Electrical Engineering and Computing, University of Zagreb \\
  Unska 3, 10000 Zagreb, Croatia \\
  \texttt{luka.pavlovic@fer.hr}}

\begin{document}
\maketitle
\begin{abstract}
In this work, we propose a method for extracting text spans that may indicate one of the BIG5 psychological traits using a question-answering task with examples that have no answer for the asked question. We utilized the RoBERTa model fine-tuned on SQuAD~2.0 dataset. The model was further fine-tuned utilizing comments from Reddit. We examined the effect of the percentage of examples with no answer in the training dataset on the overall performance. The results obtained in this study are in line with the SQuAD~2.0 benchmark and present a good baseline for further research. 
\end{abstract}

\section{Introduction}

The ability to predict psychological traits from written text without direct communication with the individual is a compelling opportunity provided by modern natural language processing methods. The recent advancements in the performance of large language models (LLMs)~(e.g. \citealp{touvron2023llama}, \citealp{chowdhery2022palm}) could further drive research in the field of automated psychological traits prediction. However, these methods do not always accurately measure what they are intended to, and advanced deep learning models may not align with psychological theory or may fail to demonstrate this alignment. This issue, known as validity, raises questions about how to improve methodology to maintain validity when using automatic reasoning in the psychological domain~\citep{Ernala2019}. In this study, we propose a method for extracting text spans that may indicate one of the BIG5 psychological traits (openness, conscientiousness, extroversion, agreeableness, and neuroticism)~\citep{costa1980influence} by utilizing a standard question-answering (QA) task to enhance validity.

\section{Methods} 
\label{section:methods}
The QA task can be solved using two distinct approaches: (1) extractive QA, and (2) generative QA. Extractive QA refers to methods that answer the question by predicting the start and end token of the candidate span within the context. Some recent models that can effectively be fine-tuned to do extractive QA are based on the Transformer architecture~\citep{https://doi.org/10.48550/arxiv.1706.03762} and include Bidirectional Encoder Representations from Transformers~(BERT)~\citep{https://doi.org/10.48550/arxiv.1810.04805}, and the more robust RoBERTa that improves on BERT by implementing careful evaluation of the effects of hyperparameter tuning and a modified pre-training task~\citep{https://doi.org/10.48550/arxiv.1907.11692}. The generative approach handles the QA problem by generating new textual sequence based on the input text rather than merely extracting the span. Models commonly used for this task include, among many other state-of-the-art models, Text-to-Text Transfer Transformer (T5)~\citep{raffel2020exploring}, and Generative Pre-trained Transformer 3 (GPT-3)~\citep{brown2020language}. Even though the generative approach has become very promising lately, it is better suited to generating answers as summaries of input rather than extracting spans in their original form. We decided on the extractive instance because we wanted to extract the span that the users wrote themselves, which we consider to be a valuable source of information.

Since we want the model to be able to give a negative answer when the trait asked for in the question is not present in the provided context, we utilized the RoBERTa model fine-tuned using the Stanford Question Answering Dataset version 2 (SQuAD 2.0)~\citep{https://doi.org/10.48550/arxiv.1806.03822}. For this work, we adopted the \texttt{roberta-base-squad2} available at Huggingface\footnote{https://huggingface.co/deepset/roberta-base-squad2}.

To investigate how the proportion of unanswerable questions in the dataset affected the final performance, we conducted three experiments with different percentages of unanswerable questions. Initially, we labeled 33\% of the dataset to be unanswerable, as originally proposed by~\citep{https://doi.org/10.48550/arxiv.1806.03822}. Secondly, we decreased the percentage to 20\% of unanswerable questions. Finally, we increased the percentage to 66\% of unanswerable questions.

\paragraph{Metrics} To evaluate the model's performance, we employed two metrics: the F1 score and the exact match (EM) metric. We utilized the EM metric to quantify the proportion of the answers that the model predicted completely correctly. However, this metric can be too strict because it considers the prediction to be incorrect even if it is one word longer or shorter than the reference. For this reason, we also apply the F1 metric to measure the similarity between the reference and the predicted answer.

We fine-tuned our model using mostly default hyperparameters. However, we increased the batch size from 8 to 16 to reduce the training time. A single fine-tuning required 13 hours on a single NVIDIA GeForce 2080 Ti graphics card. We employed the early stopping strategy based on the best results for the overall F1 score on the evaluation set.

\section{Dataset}
The dataset for this study is composed of entries following the format outlined in~\citep{https://doi.org/10.48550/arxiv.1806.03822}. The dataset fields include: (1) context, (2) question, and (3) answer which includes answer start and text.
The answer field contains the text of the correct answer and its starting character position in the context. Since some questions may have multiple correct answers for a given context, the training set had to be formatted differently from the validation set to accommodate the prescribed dataset format. In the training set, every answer to a single question for a given context was considered to be a separate data entry. In contrast, all answers for a given context and question were gathered in an array for the validation set.

The context field of our dataset was obtained from~\citep{gjurkovic-etal-2021-pandora} which comprises 17M comments from Reddit. We employed only 100k comments for training and 20k for validation. The questions were handcrafted and their format was as follows: '\texttt{What points towards psychological trait \{openness, conscientiousness, extroversion, agreeableness, neuroticism\}?}'. To obtain the answers, we extract sentences that are similar to one of the eTRSes~\citep{Gjurkovic2022}. We used cosine similarity as the metric for similarity. The similarity threshold to consider a sentence to be an answer was set at $0.63$. The trait name injected into the question corresponds to the label of the eTRS matching the sentence. 

As we mentioned in Section \ref{section:methods}, the model must learn that some questions do not have an answer for a given context. For training, a fixed portion of the answers was replaced with empty sequences to provide so-called negative examples. The question for the corresponding data entry was formed using a trait that was not found to be present in the relevant context when considering cosine similarity. We utilized this to control the proportion of these unanswerable questions in the training set. This was achieved by randomly selecting a portion of the training dataset and replacing it with negative examples while retaining the original contexts. For the validation set, we simply added the question with the trait not present in the respective context alongside the empty answer. The resulting dataset contained 126752 examples for training and 99663 for validation, of which 73071 were unanswerable.

\section{Results and Discussion} 

\begin{table*}[t]
\centering
\begin{tabular}{@{}lcccc@{}}
\toprule
             & \multicolumn{3}{c}{\% of unanswerable examples} &           \\ \cmidrule(lr){2-4}
Metric       & 20             & 33             & 66            & benchmark \\ \midrule
exact        & 73.67          & 79.83          & 84.70         & 79.87     \\
f1           & 76.93          & 83.05          & 87.32         & 82.91     \\
HasAns exact & 67.85          & 71.47          & 62.18         & 77.94     \\
HasAns f1    & 80.22          & 83.69          & 72.11         & 84.03     \\
NoAns exact  & 75.75          & 82.82          & 92.75         & 81.80     \\
NoAns f1     & 75.75          & 82.82          & 92.75         & 81.80     \\ \bottomrule
\end{tabular}
\caption{Results for different percentages of unanswerable examples. The column ``benchmark'' refers to the benchmark results from the original work 
\label{table1}
\citep{https://doi.org/10.48550/arxiv.1806.03822}. ``HasAns'' and ``NoAns'' denote the results obtained from the portion of the dataset that contains only those entries that have answers and the portion of the dataset that contains only those entries that do not contain answers, respectively.}
\end{table*}

The results of the experiments can be seen in Table~\ref{table1}. We can observe that, for all three percentages of the unanswerable examples in the train set, the results are comparable to those of the SQuAD 2.0 benchmark. The apparent improvement of the results for experimental settings with 33\% and 66\% unanswerable examples, in comparison to the benchmark, can be deceiving because those two settings obtain notably higher performance for examples with no answer while lacking in performance when considering examples that do have an answer. This, in conjunction with the fact that the support in the validation dataset is heavily out of balance in favour of the unanswerable part, attributes to the high overall performance. Due to this anomaly, the best variation for this task is probably the one with 33\% unanswerable examples in the training set which is in line with the proposed value~\citep{https://doi.org/10.48550/arxiv.1806.03822}. Another notable remark is the decline in the performance for answerable questions when their percentage is increased in the train set from 66\% to 80\%. Even though we do not know the exact reason for this phenomenon, we can speculate that this may be due to the difficulty of the underlying task of extracting psychologically indicative spans or that the fact that the model we utilized was fine-tuned using 33\% of questions without answers. 

\section{Conclusions}
In this work, we presented a novel method for extracting trait-indicative sentences from Reddit comments. We achieved this using the standard extractive QA task version that includes examples that do not contain answers in the context for a given question. To explore the importance of the percentage of the examples that do not have an answer in the train set, we did a study with three possible percentages. We obtained the best F1 score of 87.32 and the EM score of 84.70. However, due to the imbalance in the validation set, we dismiss this result in favor of the one that has more balance in the performance on both unanswerable and answerable portions of the validation set. In this case, the F1 score is 79.83, and the EM score is 83.05. These results are in line with the benchmark and we believe  this is a good baseline for further research. 

\section*{Limitations}
Please note that the experiments were run only once and the results could be affected by undetected biases in the data or by the random states within the used frameworks. Secondly, the data used for this research is not publicly available and can be requested by contacting the authors of the dataset.\footnote{\url{takelab@fer.hr}}

\section*{Acknowledgements}
This work is a result of the Research Seminar (223006) course at the Faculty of Electrical Engineering and Computing, University of Zagreb under the supervision of Prof Jan Šnajder and Matej Gjurković, MSc.

Special thanks to the Text Analysis and Knowledge Engineering Lab (TakeLab) of the Faculty of Electrical Engineering and Computing for providing computational power and the data for this research.

\bibliography{anthology,custom}
\bibliographystyle{acl_natbib}

\end{document}